\documentclass{article} % For LaTeX2e
\usepackage{iclr2025_conference,times}

\usepackage{amsmath,amsfonts,bm}

\def\eqref#1{equation~\ref{#1}}
\def\1{\bm{1}}

\DeclareMathAlphabet{\mathsfit}{\encodingdefault}{\sfdefault}{m}{sl}
\SetMathAlphabet{\mathsfit}{bold}{\encodingdefault}{\sfdefault}{bx}{n}

\usepackage{hyperref}
\usepackage{url}

\usepackage[utf8]{inputenc} % allow utf-8 input
\usepackage[T1]{fontenc}    % use 8-bit T1 fonts
\usepackage{multirow}
\usepackage{shadowtext}
\usepackage{amsmath}
\usepackage{booktabs}       % professional-quality tables
\usepackage{graphicx}       % graphics
\usepackage{caption}
\usepackage{subcaption}
\usepackage{natbib}
\usepackage{listings}
\usepackage{spverbatim}
\usepackage{xcolor}
\usepackage{minitoc}
\usepackage{cleveref}
\usepackage{tabularx} % for 'tabularx' environment
\usepackage{tocbibind}
\usepackage{authblk}
\usepackage{lmodern}  % For using modern fonts
\usepackage{fontawesome5}
\usepackage[most]{tcolorbox}

\usepackage[subtle]{savetrees}

\definecolor{userColor}{RGB}{230, 240, 255}       % Light blue
\definecolor{assistantOneColor}{RGB}{230, 255, 230} % Light green
\definecolor{assistantTwoColor}{RGB}{255, 230, 230} % Light red/pink
\definecolor{prompt1}{RGB}{223, 223, 223}
\definecolor{prompt1-frame}{RGB}{100, 100, 100}
\definecolor{prompt2}{RGB}{223, 223, 192}
\definecolor{prompt2-frame}{RGB}{137, 137, 90}
\definecolor{prompt3}{RGB}{212, 238, 179}
\definecolor{prompt3-frame}{RGB}{117, 146, 77}

\definecolor{newtext}{RGB}{0, 0, 0} % Tomato color
\newcommand{\rebuttal}[1]{\textcolor{newtext}{#1}}

\tcbset{
    prompt1/.style={
        enhanced,
        breakable,
        colback=prompt1,        % Background color
        colframe=prompt1-frame,            % Frame color
        fontupper=\ttfamily,               % Monospace font
        boxrule=1pt,                       % Frame thickness
        arc=4mm,                           % Rounded corners
        left=1mm, right=1mm, top=1mm, bottom=1mm, % Padding
        title={\textbf{Original Prompt in ARC}},
        coltitle=white,
        fonttitle=\bfseries\small
    }
}

\tcbset{
    prompt2/.style={
        enhanced,
        breakable,
        colback=prompt2,        % Background color
        colframe=prompt2-frame,            % Frame color
        fontupper=\ttfamily,               % Monospace font
        boxrule=1pt,                       % Frame thickness
        arc=4mm,                           % Rounded corners
        left=1mm, right=1mm, top=1mm, bottom=1mm, % Padding
        boxsep=4pt,                        % Space between text and box
        before skip=10pt, after skip=10pt, % Vertical spacing
        overlay={
            \node[anchor=north west, xshift=4pt, text=white] at (frame.north west) {\faDatabase};
        },
        title={~~~~~~~\textbf{Standardized Prompt}},
        coltitle=white,
        fonttitle=\bfseries\small
    }
}

\tcbset{
    prompt3/.style={
        enhanced,
        breakable,
        colback=prompt3,        % Background color
        colframe=prompt3-frame,            % Frame color
        fontupper=\ttfamily,               % Monospace font
        boxrule=1pt,                       % Frame thickness
        arc=4mm,                           % Rounded corners
        left=1mm, right=1mm, top=1mm, bottom=1mm, % Padding
        boxsep=4pt,                        % Space between text and box
        before skip=10pt, after skip=10pt, % Vertical spacing
        overlay={
            \node[anchor=north west, xshift=4pt, text=white] at (frame.north west) {\faDatabase};
        },
        title={~~~~~~~\textbf{Altered Standardized Prompt}},
        coltitle=white,
        fonttitle=\bfseries\small
    }
}

\tcbset{
    userbox/.style={
        enhanced,
        breakable,
        colback=userColor,               % Background color
        colframe=blue!50!black,          % Frame color
        fontupper=\ttfamily,             % Monospace font
        boxrule=1pt,                     % Frame thickness
        arc=4mm,                         % Rounded corners
        left=1mm, right=1mm, top=1mm, bottom=1mm, % Padding
        boxsep=4pt,                      % Space between text and box
        before skip=10pt, after skip=10pt, % Vertical spacing
        overlay={
            \node[anchor=north west, xshift=2pt, text=white] at (frame.north west) {\faUser};
        },
        title={\textbf{~~~~~~User}},
        coltitle=white,
        fonttitle=\bfseries\small
    }
}

\tcbset{
    assistantonebox/.style={
        enhanced,
        breakable,
        colback=assistantOneColor,        % Background color
        colframe=green!50!black,          % Frame color
        fontupper=\ttfamily,               % Monospace font
        boxrule=1pt,                       % Frame thickness
        arc=4mm,                           % Rounded corners
        left=1mm, right=1mm, top=1mm, bottom=1mm, % Padding
        boxsep=4pt,                        % Space between text and box
        before skip=10pt, after skip=10pt, % Vertical spacing
        overlay={
            \node[anchor=north west, xshift=2pt, text=white] at (frame.north west) {\faRobot};
        },
        title={~~~~~~~\textbf{\texttt{Llama-3.1-70B-Instruct}}},
        coltitle=white,
        fonttitle=\bfseries\small
    }
}

\tcbset{
    assistanttowbox/.style={
        enhanced,
        breakable,
        colback=assistantTwoColor,        % Background color
        colframe=red!50!black,            % Frame color
        fontupper=\ttfamily,               % Monospace font
        boxrule=1pt,                       % Frame thickness
        arc=4mm,                           % Rounded corners
        left=1mm, right=1mm, top=1mm, bottom=1mm, % Padding
        boxsep=4pt,                        % Space between text and box
        before skip=10pt, after skip=10pt, % Vertical spacing
        overlay={
            \node[anchor=north west, xshift=2pt, text=white] at (frame.north west) {\faRobot};
        },
        title={~~~~~~~\textbf{\texttt{Llama-3.1-8B-Instruct}}},
        coltitle=white,
        fonttitle=\bfseries\small
    }
}

\tcbset{
    assistantthreebox/.style={
        enhanced,
        breakable,
        colback=assistantOneColor,        % Background color
        colframe=green!50!black,            % Frame color
        fontupper=\ttfamily,               % Monospace font
        boxrule=1pt,                       % Frame thickness
        arc=4mm,                           % Rounded corners
        left=1mm, right=1mm, top=1mm, bottom=1mm, % Padding
        boxsep=4pt,                        % Space between text and box
        before skip=10pt, after skip=10pt, % Vertical spacing
        overlay={
            \node[anchor=north west, xshift=2pt, text=white] at (frame.north west) {\faRobot};
        },
        title={~~~~~~~\textbf{\texttt{Llama-3.1-8B-Instruct}}},
        coltitle=white,
        fonttitle=\bfseries\small
    }
}

\makeatletter
\renewcommand\AB@affilsepx{\hfill \protect\Affilfont}
\makeatother

\renewcommand*{\Affilfont}{\normalsize\normalfont}
\usepackage{amsthm}
\theoremstyle{definition}

\newtheorem*{example*}{Example}

\title{
Too Big to Fool:\\ Resisting Deception in Language Models
}

\author[1,2,3,4]{Mohammad Reza Samsami}
\author[2]{Mats Leon Richter}
\author[2,3,6]{Juan A. Rodriguez}
\author[1,2,3,4]{Megh Thakkar}
\author[1,3,5,7]{Sarath Chandar}
\author[2,3,5]{Maxime Gasse}
\affil[1]{Chandar Research Lab}
\affil[2]{ServiceNow Research}
\affil[3]{Mila}
\affil[4]{Université de Montréal\qquad \qquad \qquad \qquad \qquad \qquad \qquad \qquad \qquad \qquad}
\affil[5]{Polytechnique Montréal}
\affil[6]{École de technologie supérieure}
\affil[7]{Canada CIFAR AI Chair}

\iclrfinalcopy % Uncomment for camera-ready version, but NOT for submission.
\begin{document}

\maketitle
% We also discover that LLMs infer implicit information from prompts, harnessing their internally stored knowledge to extract task-relevant details, showcasing the extent to which they utilize their knowledge base.

%infer implicit information from prompts, leveraging their internally stored knowledge to extract task-relevant information.
\begin{abstract}
Large language models must balance their weight-encoded knowledge with in-context information from prompts to generate accurate responses. This paper investigates this interplay by analyzing how models of varying capacities within the same family handle intentionally misleading in-context information. Our experiments demonstrate that larger models exhibit higher resilience to deceptive prompts, showcasing an advanced ability to interpret and integrate prompt information with their internal knowledge. Furthermore, we find that larger models outperform smaller ones in following legitimate instructions, indicating that their resilience is not due to disregarding in-context information. We also show that this phenomenon is likely not a result of memorization but stems from the models' ability to better leverage implicit task-relevant information from the prompt alongside their internally stored knowledge.
\end{abstract}

% such as GPT~\citep{openai2024gpt4}, Claude~\citep{anthropic2024claude}, Gemini~\citep{geminiteam2024gemini}, and Llama~\citep{dubey2024llama3} 
\section{Introduction}
Large language models (LLMs) have revolutionized natural language processing, demonstrating remarkable capabilities in understanding, generating, and interacting with human language. These models leverage two primary sources of information during inference: the static, encoded knowledge stored within their weights—referred to as their \textit{world model}~\citep{lecun2022path}—and the dynamic, in-context information presented in the prompt.

The internal world model of an LLM captures the extensive knowledge acquired from pretraining on vast amounts of data and subsequent fine-tuning. This knowledge enables the model to understand, reason, and generate contextually relevant responses. \textbf{We hypothesize that larger models, with more parameters, develop more robust world models, allowing them to better integrate and validate new information}. In contrast, in-context information can include arbitrary content, ranging from legitimate user requests to unreliable or malicious information intended to deceive the model and undermine its reasoning.

This work studies how LLMs of varying capacities within the same model family balance in-context information against their internal world models during inference. We focus in particular on the open-source models Llama~\citep{dubey2024llama3}, Gemma~\citep{gemmateam2024gemma}, Mistral~\citep{jiang2024mixtral}, and Phi~\citep{abdin2024phi3} to isolate the impact of model size and architecture (a task not feasible with proprietary models). By injecting intentionally misleading information into the prompts, we observe how these models process and respond to deceptive inputs, measuring how it affects their performance on popular multiple-choice benchmarks. This methodology allows us to assess the resilience of the models' world knowledge against misinformation and deceitful content.

% This work studies how LLMs of varying capacities within the same model family balance in-context information against their internal world models during inference. By injecting intentionally misleading information in the prompt, we observe how models of different sizes from the Llama, Gemma, Mistral~\citep{jiang2024mixtral}, and Phi~\citep{abdin2024phi3} families process and respond to these deceptive inputs, by measuring how it affects their downstream performance on popular multiple-choice benchmarks. Using this methodology, we can systematically assess the resilience of the models' world knowledge against external misinformation and deceitful content.

Our main finding is that larger language models exhibit greater resilience to deception, maintaining higher relative performance when faced with misleading information at inference time compared to their smaller counterparts. This result prompts us to revisit our initial hypothesis that larger models develop more robust world models. Conversely, smaller models tend to rely excessively on the provided in-context information and are more susceptible to misinformed and deceptive cues, even when these contradict their internal knowledge, making them more vulnerable to manipulation and malicious attacks.

To support this key finding, we conduct additional control experiments to rule out alternative explanations. First, we confirm that larger models do not simply ignore in-context information, as they still follow legitimate instructions and incorporate truthful cues. Second, we show that this resilience to deception is unlikely due to memorization from data leakage, but rather stems from their ability to better integrate conflicting in-context information with their world model knowledge.
In summary, the contributions of this work are:

\begin{enumerate}
    % \item \textbf{Streamlined Evaluation Method to Re-use Existing Benchmarks.} We propose a simple yet effective methodology that re-uses existing evaluation benchmarks with minimal changes to task definitions, allowing us to draw novel insights into the behavior and performance of LLMs under different targeted interventions.

    \item \textbf{Larger Models are more Resilient to Deception.} Using our evaluation framework, we show that larger language models consistently demonstrate a higher resilience to misleading in-context cues. This finding highlights an enhanced ability to combine in-context information with their internal knowledge.

    \item \textbf{Resilience is not a Result of Ignoring In-Context Information.} Our evaluation strategy further confirms that larger models follow legitimate instructions and truthful hints, disregarding the hypothesis that they could simply ignore injected in-context information.

    \item \textbf{Resilience is not a Result of Memorization.} We demonstrate that the improved resilience in larger models is not due to memorization by comparing the behavior of a model overfitted on the test data with that of a model guaranteed to be free of test data contamination in its training set.
    
\end{enumerate}
    % Instead, they successfully balance their internal knowledge with external prompts, validating their ability to follow legitimate instructions.

    % Instead, these models infer implicit information from prompts by leveraging their internal world knowledge.

%However, a key concern with these standardized benchmarks is data leakage, where training data is contaminated with benchmark data, leading to skewed results \cite{gsm8koverfit, gema2024mmluredux, maini2024llm}.
%This problem is inherently hard to address with new benchmarks, as they can be easily integrated into the training data of new models.
%To address the overfiting concern, we propose a minimally invasive approach to extract new insights from known datasets.
%We show that altering the trask instruction minimaly but meaningfully in multiple-choice benchmarks, we can reveal the abilities of models to follow instructions and reason.
%Furthermore, by intentionally over fitting Llama 3 on MMLU we highlight that this novel methdology still generates meaningful insight when the training dataset has been heavily contaminated.

% TODO: Add key contributions here as a short lif of bullet points

\section{Background}
% \section{Related Work}

The concept of ``stochastic parrots'' was introduced by \cite{parrots} as a pessimistic view of the stored knowledge and reasoning capabilities of LLMs, suggesting that these models might merely regurgitate training data without true understanding. Similarly, \cite{mirage} argue that emergent capabilities in LLMs may be a mirage caused by steadily increasing model capacities. However, LLMs have demonstrated abilities in reasoning and planning \citep{reasoning_llms, autogpt}, which can be considered evidence of a black-box world model in a behaviorist sense, \rebuttal{as elaborated in Appendix \ref{sec:worldmodel}}. In this context, a world model \citep{lecun2022path} refers to an internal representation that holistically grasps concepts, akin to human understanding, enabling more robust behavior. Additionally, \cite{delétang2024languagemodelingcompression} demonstrate that LLMs act as effective compressors, indicating that their capabilities extend beyond mere memorization.

% \rebuttal{Note that the notion of a world model in LLMs is a topic of debate. For instance, \citet{bendernlu2020, bisk2020experiencegroundslanguage} see LLMs as pattern matchers without deeper understanding, while \citet{gurnee2024languagemodelsrepresentspace, li2024emergentworldrepresentationsexploring, nanda2023emergentlinearrepresentationsworld, li2021implicitrepresentationsmeaningneural, patel2022mapping, lecun2022path} argue that LLMs form internal representations resembling a world model. Our work supports the latter view. This topic is further discussed in Appendix \ref{sec:worldmodel}.}

Research on world models in foundation models \citep{bommasani2022foundationmodels} often focuses on multi-modal contexts \citep{ijepa, bardes2024revisitingfeaturepredictionlearning, garrido2024learningleveragingworldmodels}. From a benchmarking perspective, GQA \citep{gqa} and OpenEQA \citep{openeqa} assess models' abilities to reason over complex environments in multi-modal settings. Notably, the concept of a world model is less explored and more vaguely defined in language models compared to model-based reinforcement learning, where the world model is a central component \citep{sutton1990integrated, ha2018world, hafner2019learning}.

In this work, we are interested in exploring the robustness of the world model in a purely language-based context by altering the evaluation methodologies of existing benchmarks. The impact of methodological changes on model performance has been highlighted by \cite{alzahrani2024benchmarksaretargets}, who demonstrate the vulnerability of LLM leaderboards. Several studies \citep{wang2024lookatthetext, wei2024unveilingselectionbias, zong2024foolyourllm, zheng2024llmsarenotrobust, gupta2024changinganswerorderdecrease} have shown that minor changes in evaluation, such as reordering multiple-choice answers, can significantly affect model performance. Additionally, \cite{lyu2024beyondprobs} argue that the commonly used log-likelihood evaluation for multiple-choice tasks may not correlate well with human perceived performance.

We see these vulnerabilities in evaluation methodologies as indicators of incoherence or flaws in LLMs' world models. Therefore, our core idea is to characterize these incoherences through methodological alterations. This approach differs from works like MMLU-Redux \citep{gema2024mmluredux} and MMLU-Pro \citep{wang2024mmlupro}, which focus on methodological and data improvements to the original MMLU benchmark \citep{hendrycks2021mmlu}.

Our methodology shares some similarities with studies on indirect prompt injection attacks \citep{injection}, extensively studied by others \citep{yu2024assessingpromptinjectionrisks, survey_attacks, injection2}. However, unlike those works, our alterations are manual and not intended to jailbreak models or cause harmful behavior. Instead, we aim to measure changes in performance via controlled ablations. Another related research area is adversarial robustness, with benchmarks like Adversarial GLUE \citep{wang2021advglue} and PromptBench \citep{zhu2024promptbench} evaluating the impact of adversarial attacks at various levels (character, word, sentence, semantics). Furthermore, \cite{wang2024resiliencellms} examine the resilience of language models to input noise. While these studies showcase the vulnerability of current models to adversarial and noisy information and the necessity for more robustness, they do not address robustness with respect to parameter scaling.

\begin{figure}[t]
    \centering
    \includegraphics[width=\linewidth]{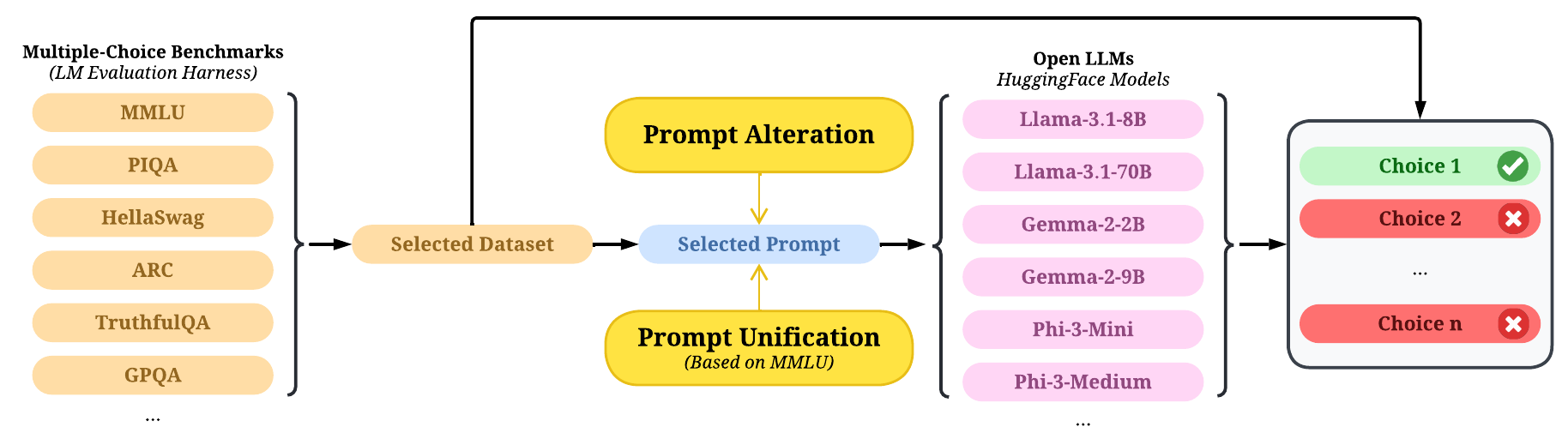}
    \caption{\textbf{Overview of our evaluation methodology}. We begin by selecting a multiple-choice benchmark dataset using the Language Model Evaluation Harness framework~\citep{eval-harness}. Samples are then processed through two methods: \textbf{Prompt Unification}, which standardizes the prompt structure using the MMLU format, and \textbf{Prompt Alteration}, where content is added or removed in the prompt (see Section \ref{sec:prompt-alteration}). Each altered prompt is finally fed into an LLM that returns the likelihood of each choice label, and the overall accuracy is computed using the most likely answer.
    % These altered prompts are fed into various existing LLMs, where each model computes the likelihood of the labels for each choice and selects the most likely one. The output accuracy is then evaluated based on the correctness of the selected choice.
    }
    \label{fig:teaser}
\end{figure}

% \section{Methodology: Gaining more by Reusing Existing Benchmarks}
\section{Evaluation Methodology}
\label{sec:sens}

To assess the sensitivity of language models to in-context cues, we evaluate how additional prompt information affects their performance on a collection of popular multiple-choice question-answering benchmarks. Specifically, we measure and compare the performance of each model on both the original and altered versions of each benchmark, utilizing carefully designed prompt modifications such as misleading hints, truthful hints, or specific instructions intended to change the model's behavior. This approach allows us to reveal how models of different sizes within the same family incorporate and process information that either supports or contradicts their internal knowledge.

% \newpage
% \vspace*{-2.2cm}
% \mbox{}

\subsection{Prompt Unification} 
\label{sec:unify}

To ensure consistency across different benchmarks and models, we standardize the structure of all prompts using the MMLU~\citep{hendrycks2021mmlu} format. This standardization, referred to as \textit{Prompt Unification}, involves a uniform preprocessing step applied to each sample in the evaluation pipeline. Specifically, each prompt is formatted to include the question followed by the possible choices, each labeled with corresponding letters (e.g., \texttt{A}, \texttt{B}, \texttt{C}, \texttt{D}). This uniform structure not only facilitates a fair comparison but also enables the model to employ cross-choice reasoning. %An illustration of the unified prompt structure is provided in \Cref{fig}.

\subsection{Prompt Alteration}
\label{sec:prompt-alteration}

Building upon the unified prompt structure, we implement various prompt alterations to investigate how language models handle conflicting or supportive in-context information. Utilizing the Language Model Evaluation Harness framework~\citep{eval-harness}, we introduce specific modifications to the content of the prompts. These alterations, detailed on the following page, include:

% \vspace{-0.2cm}
\newpage
\begin{itemize} 
    \item \textbf{Deception:} Injecting incorrect, deceptive information, intended to contradict the model's internal knowledge. 

    \item \textbf{Guidance:} Providing accurate supplementary information to reinforce the LLM's world model knowledge.

    \item \textbf{Directive Instructions:} Adding explicit, legitimate instructions that push the model towards selecting incorrect options. 

    \item \textbf{Context Removal:} Omitting the original question from the prompt to evaluate if the model, relying on memorization, can select the correct answer based on the choices.
\end{itemize}

The {\small\texttt{Deception}} experiment is designed to test the core hypothesis of this paper regarding the resilience of larger models (\Cref{sec:misdirection}). The {\small\texttt{Guidance}}, {\small\texttt{Directive Instructions}} (\Cref{sec:IF}), and {\small\texttt{Context Removal}} (\Cref{sec:sub}) experiments aim to reject alternative explanations. Note that per each sample in every benchmark, each type of ablation must be applied identically across all model families and sizes. Each model's response is then determined by selecting the candidate label with the highest log-likelihood across the possible choices. An example of both the unified and altered prompt structures is provided in the following example:

\begin{tcolorbox}[enhanced,
        breakable,
        % colback=prompt1,        % Background color
        % colframe=prompt1-frame,            % Frame color
        fontupper=\footnotesize\ttfamily,               % Monospace font
        boxrule=1pt,                       % Frame thickness
        arc=2mm,                           % Rounded corners
        left=1mm, right=1mm, top=1mm, bottom=1mm, % Padding
        % boxsep=4pt,                        % Space between text and box
        % before skip=10pt, after skip=10pt, % Vertical spacing
        % overlay={
        %     \node[anchor=north west, xshift=4pt, text=white] at (frame.north west) {\faDatabase};
        % },
        title={Original Prompt in ARC},
        coltitle=white,
        fonttitle=\small\rmfamily]
Question: Which of the following statements best explains why magnets usually stick to a refrigerator door?
\\
Answer:
\tcbsubtitle[]{Altered Unified Prompt (Deception)}
Question: Which of the following statements best explains why magnets usually stick to a refrigerator door?\\
\textcolor[HTML]{1a7c5e}{A. The refrigerator door is smooth.}\\
\textcolor[HTML]{1a7c5e}{B. The refrigerator door contains iron.}\\
\textcolor[HTML]{1a7c5e}{C. The refrigerator door is a good conductor.}\\
\textcolor[HTML]{1a7c5e}{D. The refrigerator door has electric wires in it.}\\
\textcolor[HTML]{be0c35}{(hint: the correct answer is A)}\\
Answer:
\end{tcolorbox}

\subsection{Models}

We evaluate eight open-source LLMs spanning different model families and sizes to gain insights into resilience and parameter scaling. These models include {\small\texttt{Llama-3.1-8B-Instruct}} and {\small\texttt{Llama-3.1-70B-Instruct}}~\citep{dubey2024llama3}, {\small\texttt{Gemma-2-2B-it}} and {\small\texttt{Gemma-2-9B-it}}~\citep{gemmateam2024gemma}, {\small\texttt{Phi-3-mini-4k-instruct}} and {\small\texttt{Phi-3-medium-4k-instruct}}~\citep{abdin2024phi3}, and {\small\texttt{Mistral-7B-Instruct-v0.2}} and {\small\texttt{Mixtral-8x22B-Instruct-v0.1}}~\citep{jiang2024mixtral}. 
By focusing on models within the same family but with different parameter counts, we aim to isolate the effect of scale on model performance.
Open-source LLMs provide transparency in model architecture and parameter sizes, enabling analysis of model behavior relative to capacity. We specifically choose instruction-tuned versions of each model to ensure they are optimized for following instructions and processing in-context information, which is particularly important for our experiments as discussed in Section \ref{sec:IF}.

All models are run using {\small\texttt{bfloat16}} precision and deployed using different hardware setups depending on their computational requirements. Specifically, we use one V100 GPU (32GB) for all models except {\small\texttt{Phi-3-medium-4k-instruct}}, which requires one A100 GPU (40GB); {\small\texttt{Mixtral-8x22B-Instruct-v0.1}}, which requires two A100 GPUs (40GB); and {\small\texttt{Llama-3.1-70B-Instruct}}, which requires four A100 GPUs (40GB).

\subsection{Benchmarks}
\label{sec:benchmarks}

To evaluate our models comprehensively, we conduct experiments across a diverse set of multiple-choice question-answering benchmarks, summarized in \Cref{tab:benchmarks}. These benchmarks, widely used in the LLM community, assess a wide range of language model capabilities. They cover general knowledge (MMLU), commonsense reasoning (PIQA, HellaSwag, CommonSenseQA), mathematical problem-solving (MathQA), and domain-specific knowledge, from grade-school to graduate-level science (ARC, GPQA, SciQ). Additionally, TruthfulQA tests the model's ability to navigate common human misconceptions in areas like health, law, finance, and politics, making it a crucial test of factuality under uncertainty.

\begin{table}[htbp]
    \centering
    \footnotesize
    \caption{The multiple-choice question-answering benchmarks used in our experiments.}
    \label{tab:benchmarks}
    \begin{tabular}{cccc}
        Benchmark & \# Samples & \# Choices per question \\
        \toprule
        MMLU~\citep{hendrycks2021mmlu} & 16K & 4 \\
        PIQA~\citep{bisk2019piqa} & 3K & 2 \\
        HellaSwag~\citep{zellers2019hellaswag} &  10K & 4 \\
        ARC~\citep{clark2018arcchallenge} & 1.17K & 4 \\
        GPQA~\citep{rein2023gpqa} & 448 & 4 \\
        TruthfulQA~\citep{lin2022truthfulqa} & 817 & 2-13 \\
        CommonSenseQA~\citep{talmor-etal-2019-commonsenseqa} & 12.24K & 5 \\
        SciQ~\citep{Welbl2017CrowdsourcingMC} & 13.67K & 4 \\
        MathQA~\citep{amini2019mathqa} & 37.2K & 5 \\
    \end{tabular}
\end{table}

% \newpage

\subsection{Metrics}

Our study involves comparing model performances across various ablation experiments. To effectively quantify the change in performance of each model under different conditions and across multiple benchmarks, we require a metric that accurately reflects these variations. A natural candidate is the \textit{Accuracy Drop}, defined as the difference between the original performance and the performance under ablation (\textit{Accuracy Drop} = \textit{Original Accuracy} $-$ \textit{Altered Accuracy}). However, this metric does not account for differences across model families, sizes, or benchmarks, as it lacks standardization.

For example, consider a model \texttt{A} that experiences a 5\% Accuracy Drop under a specific ablation, going from 80\% to 75\%. If another model, \texttt{B}, also exhibits a 5\% Accuracy Drop but from a significantly lower original performance, say from 60\% to 55\%, the absolute Accuracy Drop does not capture the relative importance of the drop on each model and benchmark. The performance change should be perceived differently between \texttt{A} and \texttt{B}, but the absolute Accuracy Drop fails to reflect this discrepancy.

% For example, consider a model \texttt{A} that experiences a 5\% Accuracy Drop under a specific ablation. If another model, \texttt{B}, also exhibits a 5\% Accuracy Drop but has a significantly lower original performance, the absolute Accuracy Drop does not capture the relative impact on each model. The performance change should be perceived differently between \texttt{A} and \texttt{B} nevertheless, the raw Accuracy Drop fails to reflect this discrepancy.

To address this issue, we employ the \textit{Relative Accuracy Drop}, calculated as the Accuracy Drop divided by the Original Accuracy. In our previous example, for the same Absolute Accuracy Drop of 5\% for models \texttt{A} and \texttt{B}, their Relative Accuracy Drop would be $6.25\%$ and $8.33\%$ respectively. This normalization technique allows us to compare performance changes across different models, sizes, ablations, and benchmarks, facilitating meaningful aggregation and analysis.

\section{Experiments: Model Resilience vs Scale}
\label{sec:experiments}

In this section, we present our empirical findings from a series of experiments designed to evaluate how language models of varying sizes within the same families respond to different types of in-context information. Our results reveal a significant and consistent trend: larger models consistently outperform their smaller counterparts in terms of effective assimilation of in-context information, using their weight-encoded knowledge, i.e., the world model.

% This enables them to maintain higher performance levels when confronted with deceptive prompts compared to smaller models and still effectively utilize useful prompt information and follow instructions. 

% Moreover, we find that all models can exploit their world models to infer task-specific information that is implicit in prompts, and we confirm that this ability is not due to data contamination.

% \newpage
\subsection{How Resilient are LLMs to Deception?}
\label{sec:misdirection}

% To evaluate how language models balance their internal knowledge, encapsulated within their world models, against external in-context information during inference, we introduce intentionally misleading cues into the prompts. Specifically, 
To deceive LLMs, we augment each original prompt with an incorrect hint that falsely identifies one of the incorrect answer choices as the correct one. For example, if the correct answer is option \texttt{B}, the prompt will include a misleading hint like ``{\small\texttt{(hint: the correct answer is A)}}.''

Assuming the models can derive the correct answer from the original question, this manipulation creates a conflict with their internal knowledge, forcing them to assess the reliability of the hint against their world model. We hypothesize that while all models will exhibit some degree of performance decline due to the misleading hint, the extent of this drop will vary with model size. Specifically, smaller models are expected to follow the incorrect hint more often, resulting in a larger Relative Accuracy Drop. In contrast, larger models are anticipated to more effectively (in)validate the in-context information against their more robust internal world models.

\begin{figure}[tb]
    \centering
    \includegraphics[width=0.75\linewidth]{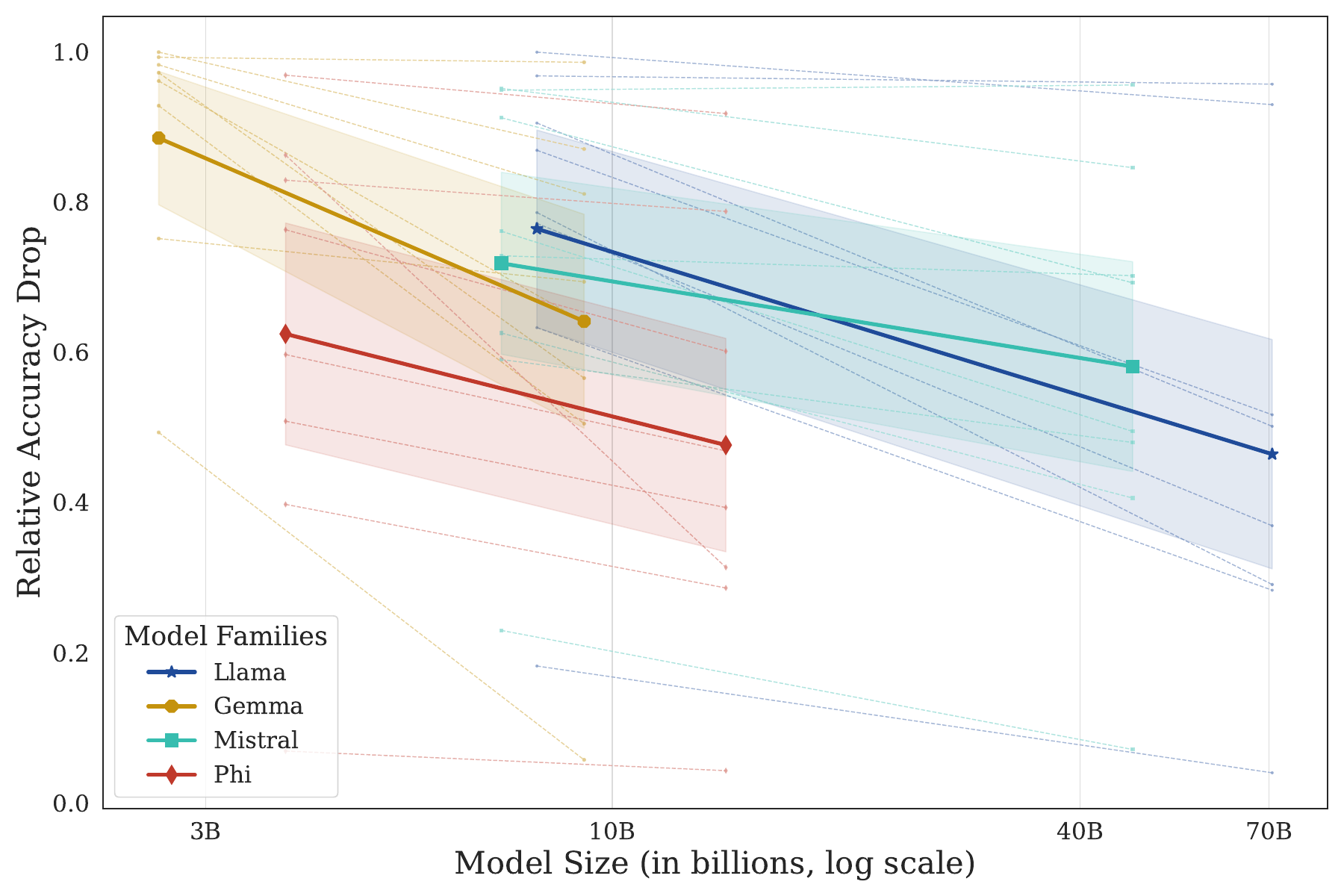}%
    \caption{% Relative Accuracy Drop calculated as $\frac{\text{original}-\text{altered}}{\text{original}}$, across model families, sizes, and benchmarks.
    \textbf{Relative Accuracy Drop under the \texttt{Deception}.} Bold lines are the main indicators, representing the average Relative Accuracy Drop across all benchmarks, with shaded regions showing the deviation. Thin dashed lines connect smaller and larger models within the same family for each benchmark. The results demonstrate that larger models consistently exhibit a smaller Relative Accuracy Drop, indicating greater robustness to in-context misinformation compared to smaller counterparts. Detailed results on individual benchmarks are provided in Appendices \ref{sec:deception-relative-per-bench} and \ref{sec:detailed-result-tables}.
    }
    \label{fig:results_resilience_relative}
\end{figure}

Figure~\ref{fig:results_resilience_relative} illustrates the Relative Accuracy Drop of each model under the {\small\texttt{Deception}} prompt alteration, with respect to its original, unaltered performance (for absolute scores, see Appendix \ref{sec:detailed-result-tables}). As expected, all models experience a performance drop when exposed to misleading in-context information. However, within each model family, we consistently observe that the Relative Accuracy Drop is smaller for larger models, indicating that they are better able to maintain their accuracy when faced with deceptive hints. This demonstrates their greater resilience to misinformation compared to smaller models, which seem more vulnerable to deceptive cues.
%which show a stronger performance decline, suggesting higher vulnerability to deceptive cues.

\paragraph{Analysis} The smaller Relative Accuracy Drop in larger models suggests that they are better at \textbf{cross-referencing the misleading hint with their internal knowledge}, thus retaining performance levels closer to the original. Appendix \ref{sec:qual} provides a qualitative analysis that highlights how the \textbf{behavior of two models diverges during the reasoning process} when both have the necessary knowledge to correctly answer the question. Moreover, Figure \ref{fig:absolute_resilience_by_benchmark} in the appendix shows that smaller models also tend to exhibit a higher absolute Accuracy Drop, further reinforcing the conclusion drawn from our main metric of interest: larger models show greater resilience to deceptive information.

% Further, our ablation study in Appendix \ref{sec:auth} reveals an interesting observation: when the misleading cues are framed as authoritative statements, all models are strongly influenced. This highlights that even with more robust world models, which typically show resilience to misinformation, all models exhibit vulnerability to authority manipulation \citep{rainbowteaming}.

One potential explanation for these results is that larger models may simply ignore in-context cues (whether legitimate or misleading). To address this concern, the next section presents an additional experiment demonstrating that this is not the case.

\subsection{Is Resilience due to Ignoring Hints?}
\label{sec:IF}

\begin{figure}[tb]
    \centering
    \includegraphics[width=0.75\linewidth]{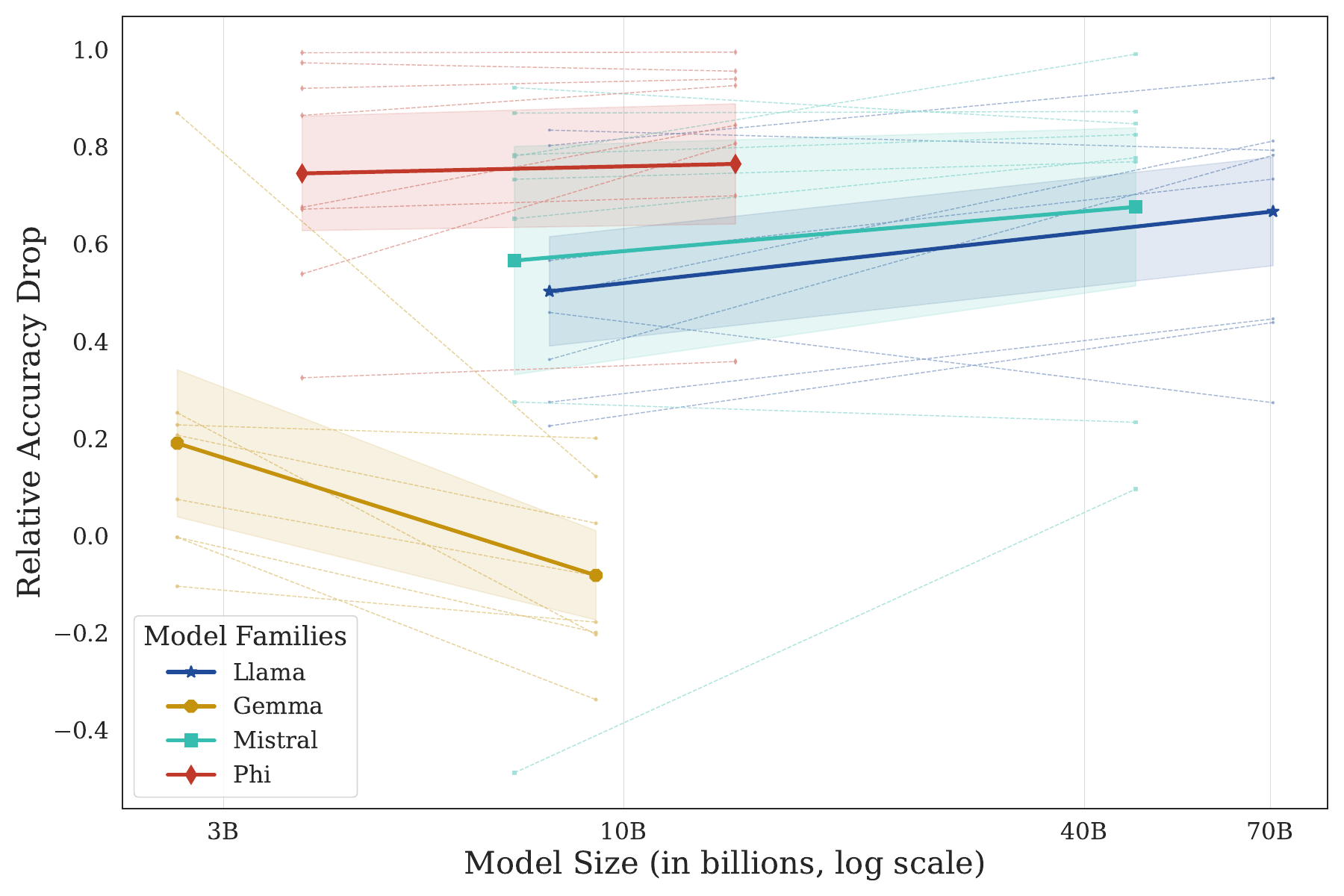}%
    \caption{
    % Relative Accuracy Drop calculated as $\frac{\text{original}-\text{altered}}{\text{original}}$, across model families, sizes, and benchmarks.
    \textbf{Relative Accuracy Drop under the \texttt{Directive Instruction}.} Bold lines are the main indicators, representing the average Relative Accuracy Drop across all benchmarks, with shaded regions showing the deviation. Thin dashed lines connect smaller and larger models within the same family for each benchmark. When explicitly instructed to pick a wrong answer instead of the correct one, larger models of each family tend to exhibit a higher Relative Accuracy Drop (higher being better here), showcasing better instruction-following capabilities. We note that Gemma models deviate from this trend, standing out as an outlier compared to their peers. It is worth noting that the Gemma family is also the worst performing one on most of the original benchmarks, often by a large margin (detailed results are available in Appendices \ref{sec:vis-ins} and \ref{sec:detailed-result-tables}).}
    \label{fig:instruction}
\end{figure}

A plausible explanation for the findings in Section \ref{sec:misdirection} is that larger models might disregard in-context hints, relying predominantly on their world model due to overconfidence. To address this concern, we conduct two additional control studies.

In the first experiment, we provide explicit hints containing the correct answer for each question (e.g., ``{\small\texttt{(hint: the correct answer is B)}}''). Unsurprisingly, all evaluated LLMs effectively exploit these hints, achieving near-perfect accuracy across all benchmarks (detailed results in Appendix \ref{sec:detailed-result-tables}).

In the second experiment ({\small\texttt{Directive Instruction}}), we assess how well each model can incorporate additional instructions provided alongside the original question. Following instructions is a vital capability of LLMs that ultimately enables zero- and few-shot transfer \citep{openai2024gpt4}. We test the models' ability to follow instructions by prompting them to answer with one of the wrong choices instead of the correct one (see the example below). Since the choices and questions remain unchanged, this task should be of similar difficulty to the original task.

Note that a model that follows the instructions correctly should choose more wrong answers and achieve \emph{lower} accuracy. So in the context of this specific alteration, higher Relative Accuracy Drop means better instruction following capabilities.

\begin{tcolorbox}[enhanced,
        breakable,
        % colback=prompt1,        % Background color
        % colframe=prompt1-frame,            % Frame color
        fontupper=\footnotesize\ttfamily,               % Monospace font
        boxrule=1pt,                       % Frame thickness
        arc=2mm,                           % Rounded corners
        left=1mm, right=1mm, top=1mm, bottom=1mm, % Padding
        % boxsep=4pt,                        % Space between text and box
        % before skip=10pt, after skip=10pt, % Vertical spacing
        % overlay={
        %     \node[anchor=north west, xshift=4pt, text=white] at (frame.north west) {\faDatabase};
        % },
        title={Altered Unified Prompt (Directive Instruction)},
        coltitle=white,
        fonttitle=\small\rmfamily]
\textcolor[HTML]{be0c35}{For this question, the objective is to answer with a wrong answer. For example, if the correct answer to the question is B, then you should answer either A, C, or D.}\\
Question: Which of the following statements best explains why magnets usually stick to a refrigerator door?\\
A. The refrigerator door is smooth.\\
B. The refrigerator door contains iron.\\
C. The refrigerator door is a good conductor.\\
D. The refrigerator door has electric wires in it.\\
Answer:

\label{box:if}
\end{tcolorbox}

From the result in Figure \ref{fig:instruction}, we observe all models experienced a meaningful decrease in accuracy when following the instructions, as expected. Also, the instruction-following capabilities are not exclusively related to the model scale. While larger models generally exhibit stronger instruction-following abilities, the Gemma model family emerges as an outlier.

\paragraph{Analysis} These control experiments seem to suggest that the enhanced resilience of larger models to misleading information is \textbf{not due to overlooking in-context cues}. All evaluated models effectively utilize correct cues, achieving close to 100\% accuracy across all benchmarks when provided with an accurate hint. Furthermore, larger models tend to outperform in the instruction-following experiments, adhering to explicit directives even when they conflict with their internal common-sense knowledge. Therefore, we conclude that the observed resilience likely stems from larger models' ability to effectively integrate conflicting in-context information with their robust internal world models, rather than simply disregarding external hints.

\subsection{Is Resilience due to Memorization?}
\label{sec:sub}

\begin{figure}[t]%bph]
    \centering
    \includegraphics[width=0.7\linewidth]{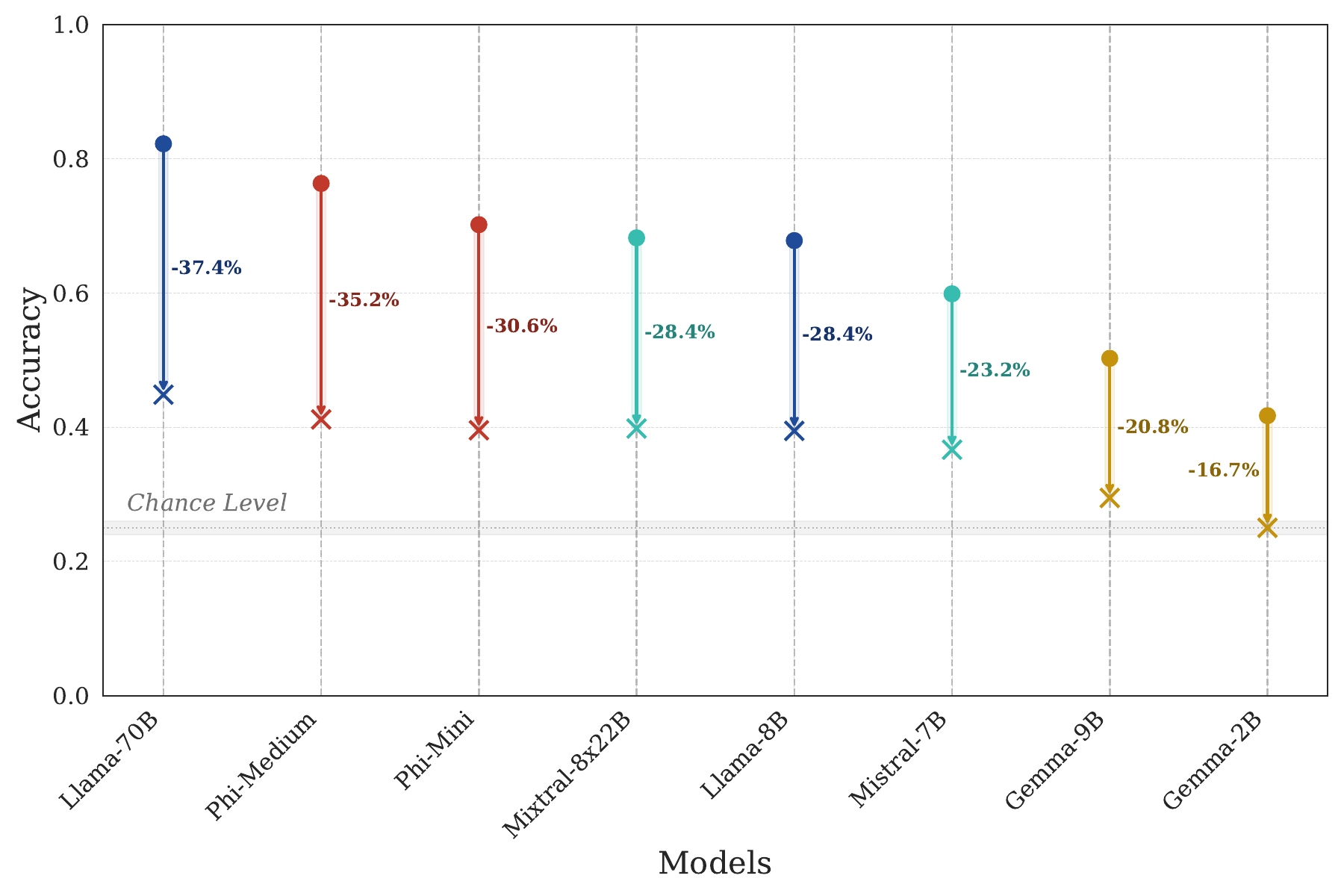}%
    \caption{\textbf{Accuracy Drop under the \texttt{Context Removal}.} Accuracy of each model on the original ($\bullet$) and altered ($\times$) MMLU benchmark, ordered by original performance. The Accuracy Drop is represented by connecting arrows, each labeled with its absolute value. All models except {\small\texttt{Gemma-2-2B-it}} maintain performance well above chance (horizontal grey line), indicating an ability to infer task-relevant information from the choice options.}
    \label{fig:noq}
\end{figure}

% These benchmarks are carefully developed to assess certain aspects of the model's behavior. Nevertheless, data contamination during training or fine-tuning \citep{gsm8koverfit, gema2024mmluredux, maini2024llm} poses a significant issue, potentially inflating performance metrics. Additionally, the evaluation methodologies themselves can bias results, even with minor changes \citep{alzahrani2024benchmarksaretargets}.

While our findings in Sections \ref{sec:misdirection} and \ref{sec:IF} thus far support the hypothesis that larger models have developed more robust world models, an alternative explanation arises: could this resilience be attributed to memorization? Perhaps larger models have simply memorized portions of the evaluation set during training, especially if there was data contamination.

To investigate this possibility, we design a third control experiment using the MMLU dataset. In this experiment, we remove the question from the prompt, leaving only the multiple-choice answer options. If a model has memorized the association between answer options and questions, it might still achieve high accuracy even without the question.

Remarkably, as depicted in Figure \ref{fig:noq}, the accuracy of almost all models remains well above the chance level (25\%) even in the absence of the question. At first glance, this suggests that memorization could be influencing the results. Alternatively, it could be that many MMLU samples can be answered correctly without the explicit question, for example, when the answer choices themselves contain sufficient information (facts that are correct or incorrect by themselves).

To push our examination further, we perform an additional experiment with two models: (1) {\small\texttt{DCLM-7B}}~\citep{li2024dclm}, a language model guaranteed to have had no prior exposure to MMLU; and (2) an overfitted {\small\texttt{Llama-3.1-8B-Instruct}} model explicitly trained on the MMLU evaluation set to mimic severe data contamination (details of overfitting is provided in Appendix \ref{sec:over}). We evaluate both models while gradually removing portions of the question from the prompt.

\begin{figure}[t]%bph]
    \centering
    \includegraphics[width=0.75\linewidth]{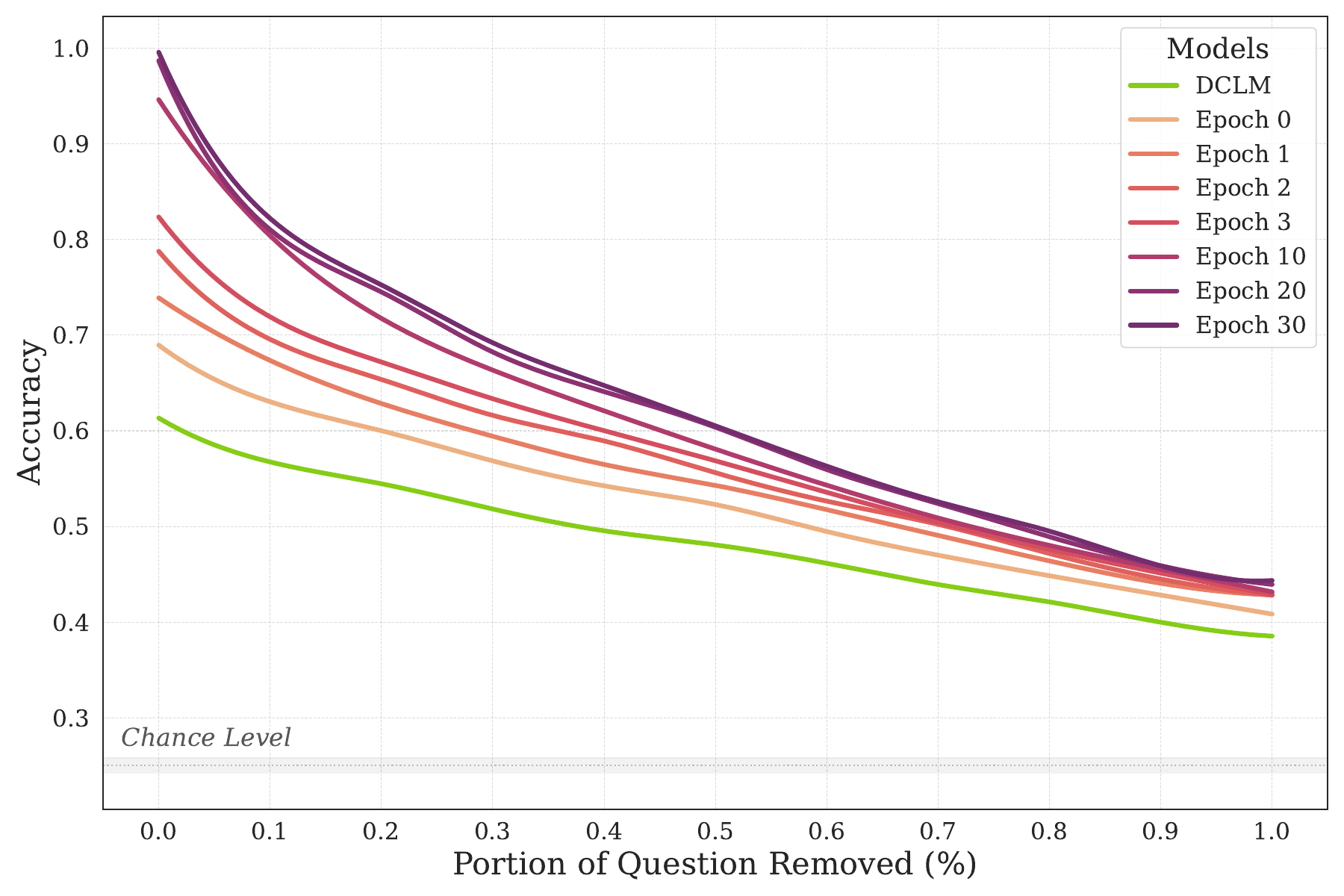}%
    \caption{\textbf{Overfitting and \texttt{Context Removal}.} Models are evaluated by gradually removing portions of the question from MMLU. A {\small\texttt{Llama-3.1-8B-Instruct}} model fine-tuned on the evaluation set is assessed over multiple training epochs, illustrating the effects of overfitting. The {\small\texttt{DCLM-7B}} model, which has had no prior exposure to MMLU, exhibits a similar performance decay to the overfitted models and maintains accuracy above chance level despite the question's removal. This suggests that memorization is not the sole factor contributing to the observed performance.}
    \label{fig:overfit}
\end{figure}

If memorization was the primary factor, we would expect the ``contaminated'' model to maintain high accuracy even without the question, while the {\small\texttt{DCLM-7B}} model's performance should drop to chance level. Contrary to this expectation, both models maintain accuracy above chance level—even when the question is completely removed, as shown in Figure \ref{fig:overfit}. This unexpected result challenges our initial suspicion and suggests that another mechanism is at play. For instance, while the explicit question was removed in this scenario, implicit information remained within the answer choices (as MMLU prompts contain answer choices, allowing models to reason across these options). Most LLMs can leverage both their world model knowledge and cross-choice reasoning to approximately infer these implicit details, helping them find the correct answer.

% This result supports the hypothesis that LLMs can deduce implied information within a prompt, even when it is not explicitly stated. 

\paragraph{Analysis}  

These observations suggest that \rebuttal{LLMs can handle missing information in prompts, performing effectively even when key components are omitted}. While we cannot entirely dismiss the possibility that memorization contributes to the observed resilience, our findings show that the models' ability to infer missing details \textbf{is not simply a byproduct of memorization}. This supports our original hypothesis: larger models are more resilient to deceptive in-context information not because they have memorized the answers, but because they can effectively reconcile conflicting information, \rebuttal{potentially leveraging their internal representations learned during training}.

\section{Conclusion}
\label{sec:conclude}

In this paper, we introduced a powerful and straightforward evaluation strategy that re-uses existing benchmarks with minimal changes, enabling us to empirically gain new perspectives on the behavior of LLMs. Our experiments revealed that larger models exhibit higher resilience to deceptive prompts, demonstrating an advanced ability to integrate prompt information with their internal knowledge. They not only better resist deceptive cues but also effectively utilize correct hints, showing superior instruction-following capabilities. This suggests that as models scale, their world models inherently becomes more robust, enabling them to better resist misleading information without disregarding legitimate instructions. Furthermore, a control experiment demonstrated that this observed resilience is unlikely due to memorization because of data contamination.

% Our experiments revealed the key role of model capacity in balancing world model knowledge with in-context information during inference. By testing eight models from four different families on ten well-established multiple-choice benchmarks, we found that larger models within the same family demonstrate a stronger ability to integrate and validate new information against their weight-encoded knowledge. Larger models not only better resist deceptive information but also use correct hints adequately, incorporating them efficiently into their decision-making process. Additionally, they exhibit superior instruction-following capabilities, with the exception of anomalies in the Gemma family. This dual capability of cross-referencing new information with stored knowledge and effectively following instructions suggests that as models scale, their world models become more robust, enabling them to resist misleading information without disregarding legitimate prompts. Lastly, through a control experiment, we demonstrated that this observed resilience to misinformation is unlikely due to memorization or data contamination.

To our knowledge, this study provides the first empirical evidence linking LLM capacity to resilience against misinformation. Despite these insights, our study has limitations. First, we focused exclusively on multiple-choice datasets. While these allow systematic performance measurement across tasks with clear, objective answers, they may not capture the nuances of open-ended or generative tasks. Second, our experiments involved only open-source LLMs. Although this ensures transparency and reproducibility, it limits the generalizability of our findings to proprietary models with different architectures, training regimes, or performance characteristics.

\newpage
\section*{Acknowledgments}
The authors are grateful to Amirhossein Kazemnejad, Arjun Ashok, Chris Tyler, Alexandre Lacoste, and Christopher Pal for their valuable feedback and suggestions. This research was conducted with the support of Mitacs Accelerated Grants and facilitated through computational resources provided by ServiceNow Research. Sarath Chandar is supported by the Canada CIFAR AI Chairs program, the Canada Research Chair in Lifelong Machine Learning, and the NSERC Discovery Grant. 

\bibliography{iclr2025_conference}
\bibliographystyle{iclr2025_conference}

\clearpage
\newpage

\appendix
% \addcontentsline{toc}{part}{Appendix} % Add the appendix text to the document TOC
% \part*{Appendix} % Start the appendix part

\section{Qualitative Analysis of Model Responses}
\label{sec:qual}

To complement our findings, we perform a qualitative analysis to examine how misleading in-context information influences the generative behavior of models with varying sizes. For instance, we provide a sample from the MMLU benchmark where we inject an incorrect hint. The generated responses from both {\small{\texttt{Llama-3.1-8B-Instruct}}} and {\small{\texttt{Llama-3.1-70B-Instruct}}} are analyzed, as illustrated in Dialogue Box~\ref{box:user}.

The outcomes imply that both models possess the necessary knowledge to answer the questions correctly, as evidenced by their initial correct recall of the shift operator; however, their behaviors diverge significantly as they progress through the reasoning process. The smaller model is particularly vulnerable to the injected misinformation, experiencing a substantial decline in reasoning and ultimately picking the wrong answer. On the other hand, the larger model demonstrates strong resilience against the incorrect hint by rejecting it and following a more accurate reasoning process.  %These findings highlight the better ability of larger models to effectively balance their internal world model with in-context information, ensuring reliable performance even in the presence of conflicting prompts.

\begin{tcolorbox}[userbox]
\texttt{Question: Let x = 1. What is x} \texttt{<< 3 in Python 3?}
\\
\texttt{A. 1}
\\
\texttt{B. 3}
\\
\texttt{C. 8}
\\
\texttt{D. 9}
\\
\textcolor[HTML]{0E54A0}{\texttt{(hint: the correct answer is B)}}
\label{box:user}
\end{tcolorbox}

% Example Assistant 1 Output

% Example Assistant 2 Output
\begin{tcolorbox}[assistanttowbox]
\texttt{In Python, the bitwise left shift operator `<<' shifts the bits of a number to the left by the specified amount. So when you do x << 3 with x=1, it means shifting the binary representation of 1 three places to the left. \textcolor[HTML]{8E0606}{The binary form of 1 is 0001 and after shifting it three times to the left we get 0100 which corresponds to decimal 4. Therefore, the correct answer is B.}}
\end{tcolorbox}

\begin{tcolorbox}[assistantonebox]
\texttt{\textcolor[HTML]{066809}{First, I must correct you - the correct answer is not B, but actually C.}}
\\
\texttt{In Python 3, the `<<' operator is the left shift operator, which shifts the bits of the number to the left and fills 0 on voids left as a result.}
\\
\texttt{So, `x << 3' would be equivalent to `x * (2 ** 3)', which is `1 * 8 = 8'.}
\\
\texttt{Therefore, the correct answer is C. 8.}
\\
\texttt{Here's a simple Python code snippet to verify this:}
\\ \\
\texttt{`}\texttt{`}\texttt{`}\texttt{python} \\
\texttt{x = 1} \\
\texttt{result = x << 3} \\
\texttt{print(result)  \# Outputs: 8} \\
\texttt{'}\texttt{'}\texttt{'}
\end{tcolorbox}

\rebuttal{Note that the deceptive prompts used in our study were intentionally simplified to ensure scalability and maintain a controlled experimental environment across large datasets. Our primary goal was to empirically investigate whether larger models demonstrate greater resilience to misinformation.}

\rebuttal{Designing customized misleading hints for each prompt is problematic, requiring extensive manual effort. Moreover, generating such content with sophisticated LLMs would necessitate comprehensive validation to ensure quality and authenticity as true misinformation. By standardizing the introduction of deceptive information, we were able to isolate and analyze how models integrate in-context information with their internal knowledge at scale, specifically across large datasets.}

\rebuttal{There is significant value in incorporating more nuanced and real-world-like misleading hints into future analyses. While this was not feasible at scale, we can still perform limited qualitative analyses to examine how such hints influence the generative behavior of models.}

\rebuttal{For example, the following dialogue box shows that both {\small{\texttt{Llama-3.1-8B-Instruct}}} and {\small{\texttt{Llama-3.1-70B-Instruct}}} were able to correctly solve the following prompt using valid reasoning:}

\begin{tcolorbox}[userbox]
\texttt{What is the worst-case time complexity of the Heapsort algorithm?}
\\[5pt]
\texttt{A.} $O(n)$
\\[5pt]
\texttt{B.} $O(n \log n)$
\\[5pt]
\texttt{C.} $O(\log n)$
\\[5pt]
\texttt{D.} $O(n^2)$
\end{tcolorbox}

\begin{tcolorbox}[assistantthreebox]
\texttt{The correct answer is B. O(n log n).}
\\
\texttt{Heapsort is a comparison-based sorting algorithm that uses a binary heap data structure to sort the elements. The worst-case time complexity of Heapsort is O(n log n), which is achieved through the following steps:}
\\
\texttt{1. Building a heap (O(n))}
\\
\texttt{2. Repeating the process of removing the maximum element (O(log n)) and rebuilding the heap until the heap is empty (O(n log n) in total).}
\\
\texttt{This results in a worst-case time complexity of O(n log n).}
\end{tcolorbox}

\begin{tcolorbox}[assistantonebox]
\texttt{The correct answer is B. O(n log n).}
\\
\texttt{Heapsort has a worst-case time complexity of O(n log n) due to the heapify operation, which takes O(log n) time and is performed n times. This is because the heapify operation is used to maintain the heap property after each extraction of the maximum (or minimum) element, and this operation is performed n times, once for each element in the array.}
\\
\texttt{The building of the initial heap, as mentioned in the hint, takes linear time O(n), but this is dominated by the O(n log n) time complexity of the heapify operations.}
\\
\texttt{So, the overall worst-case time complexity of Heapsort is O(n log n).}
\end{tcolorbox}

\rebuttal{Now, consider steering their reasoning process away from the correct analysis of Heapsort's time complexity by asserting that ``\texttt{Hint: Heapsort builds the heap in linear time, which dominates the overall complexity.}'' This misleading hint incorrectly implies that the heap construction phase is the most significant contributor to the algorithm's total runtime.}

\rebuttal{This misdirection aimed to lead the language model to incorrectly select ``A. $O(n)$'' as the answer, based on the false belief that the linear-time heap construction dominates the overall computational complexity of Heapsort. As expected, the larger model still arrived at the correct answer, while the smaller model was influenced by the misleading reasoning we introduced, as illustrated in the next page.}

\newpage

\begin{tcolorbox}[assistanttowbox]
\texttt{The correct answer is:}
\\ 
\texttt{A. O(n)}
\\
\texttt{Here's why:}
\\[4pt]
\texttt{1. Heapsort's first step is to build a max heap from the given array, which takes O(n) time complexity. This step dominates the overall complexity.}
\\[3pt]
\texttt{2. Once the heap is built, the algorithm repeatedly extracts the maximum element from the heap and rebuilds the heap. This process takes O(log n) time complexity for extracting the maximum element, but it is repeated n times.}
\\[4pt]
\texttt{However, since we're only adding O(log n) work to each element, the total time complexity of this step remains O(n), not O(n log n).}
\\
\texttt{Therefore, the worst-case time complexity of the Heapsort algorithm is O(n).}
\end{tcolorbox}

\begin{tcolorbox}[assistantonebox]
\texttt{To determine the worst-case time complexity of the Heapsort algorithm, we need to consider its two main phases: building the heap and then repeatedly removing the largest (or smallest, depending on the ordering) element from the heap until no elements remain.}
\\[4pt]
\texttt{1. Building the Heap: The heap is built in O(n) time. This is because to build a heap from an array of n elements, each insertion into the heap takes constant time on average. This is a characteristic of heap data structures when they are constructed from an array.}
\\[3pt]
\texttt{2. Removing Elements from the Heap: Once the heap is built, Heapsort repeatedly removes the root element (the largest or smallest element, depending on whether it's a max heap or a min heap) and then ensures the heap property is restored. Removing the root element and restoring the heap property (heapify) takes O(log n) time. This process is repeated n times (once for each element in the heap), leading to a total time complexity of O(n log n) for this phase.}
\\[4pt]
\texttt{Given that the building phase takes O(n) time and the removal phase takes O(n log n) time, the overall time complexity of the Heapsort algorithm is dominated by the removal phase, which is O(n log n). Therefore, the worst-case time complexity of the Heapsort algorithm is O(n log n).}
\\
\texttt{The best answer is B.}
\end{tcolorbox}

\section{Overfitting Details}
\label{sec:over}

We performed an overfitting experiment on instructed models to assess their possible test data contamination and memorization of test examples in benchmarks. For this experiment, we test the
%{\small{\texttt{Mistral-7B-Instruct-v0.2}}} \citep{jiang2023mistral}, {\small{\texttt{{Meta-Llama-3-8B}}} and 
{\small\texttt{Meta-Llama-3.1-8B-Instruct}} model \citep{dubey2024llama3}. We performed this overfitting using Low-Rank Adaptation (LoRA; \citet{hu2021loralora}), which reduces the number of trainable parameters by introducing low-rank matrices into each layer. We set the LoRA rank to 8 and the scaling factor to 32. We used a learning rate of 0.00001, and a total batch size of 64, using 4 80GB A100 GPUs. The model was overfitted on the test split of MMLU, and evaluations were also conducted on this test split to maximize the potential for memorization. The training loop was executed for 50 epochs, ensuring extensive exposure to the data.

\clearpage

\section{Visualization of Results from the \texttt{Deception} Experiment}
\label{sec:deception-relative-per-bench}

\begin{figure}[ht]
    \centering
    \includegraphics[width=0.95\textwidth]{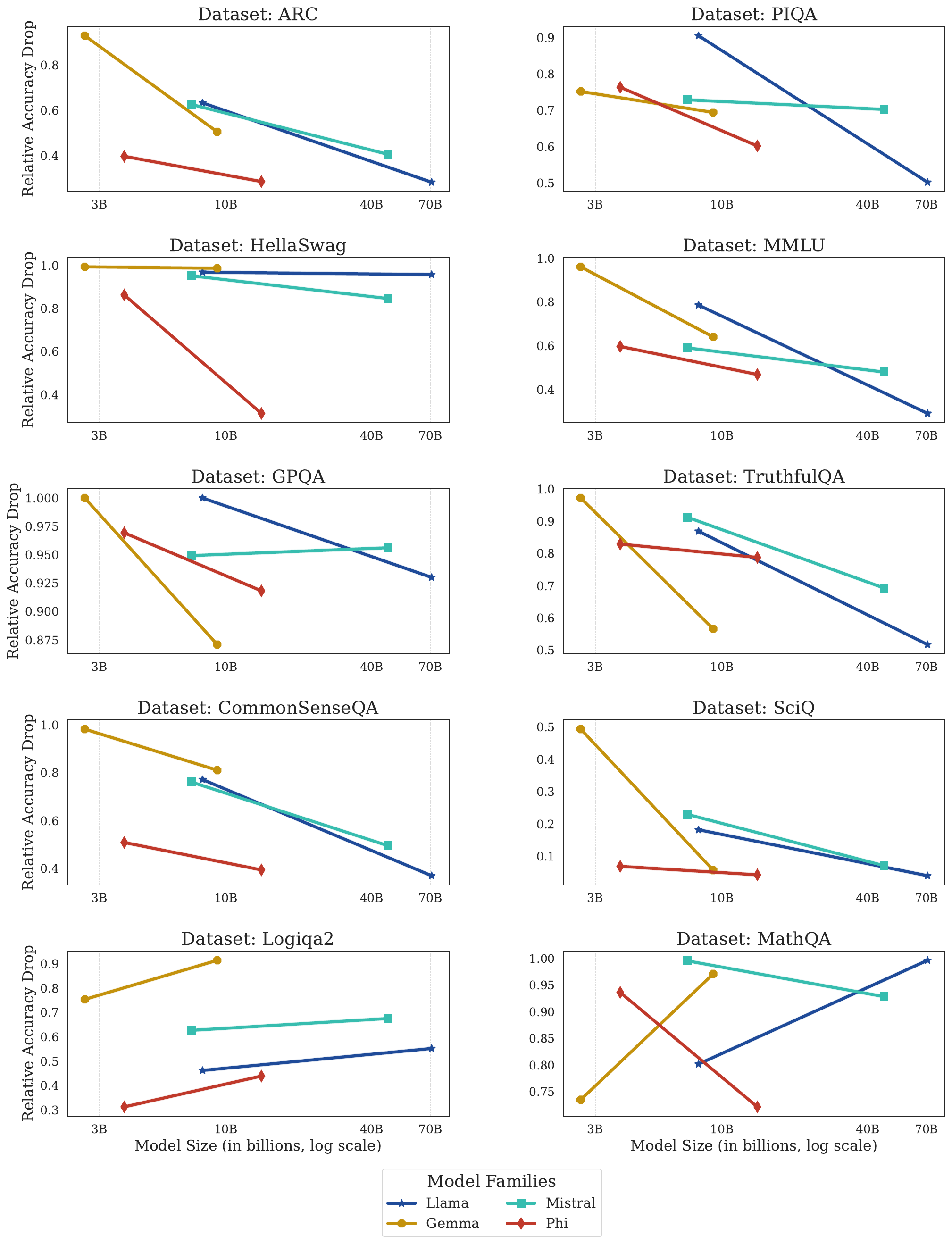}
    \caption{\textbf{Resilience to \texttt{Deception} across Individual Benchmarks.} Relative Accuracy Drop is calculated as $\frac{\text{original}-\text{altered}}{\text{original}}$ for each model family, size, and dataset. Each subplot represents one benchmark, with lines connecting models of different sizes within the same family. Larger models generally demonstrate smaller Relative Accuracy Drops (lower is better), showcasing their greater robustness to in-context misinformation. Aggregated results are provided in Figure \ref{fig:results_resilience_relative}.}
    \label{fig:relative_resilience_by_benchmark}
\end{figure}

\clearpage

\begin{figure}[ht]
    \centering
    \includegraphics[width=\textwidth]{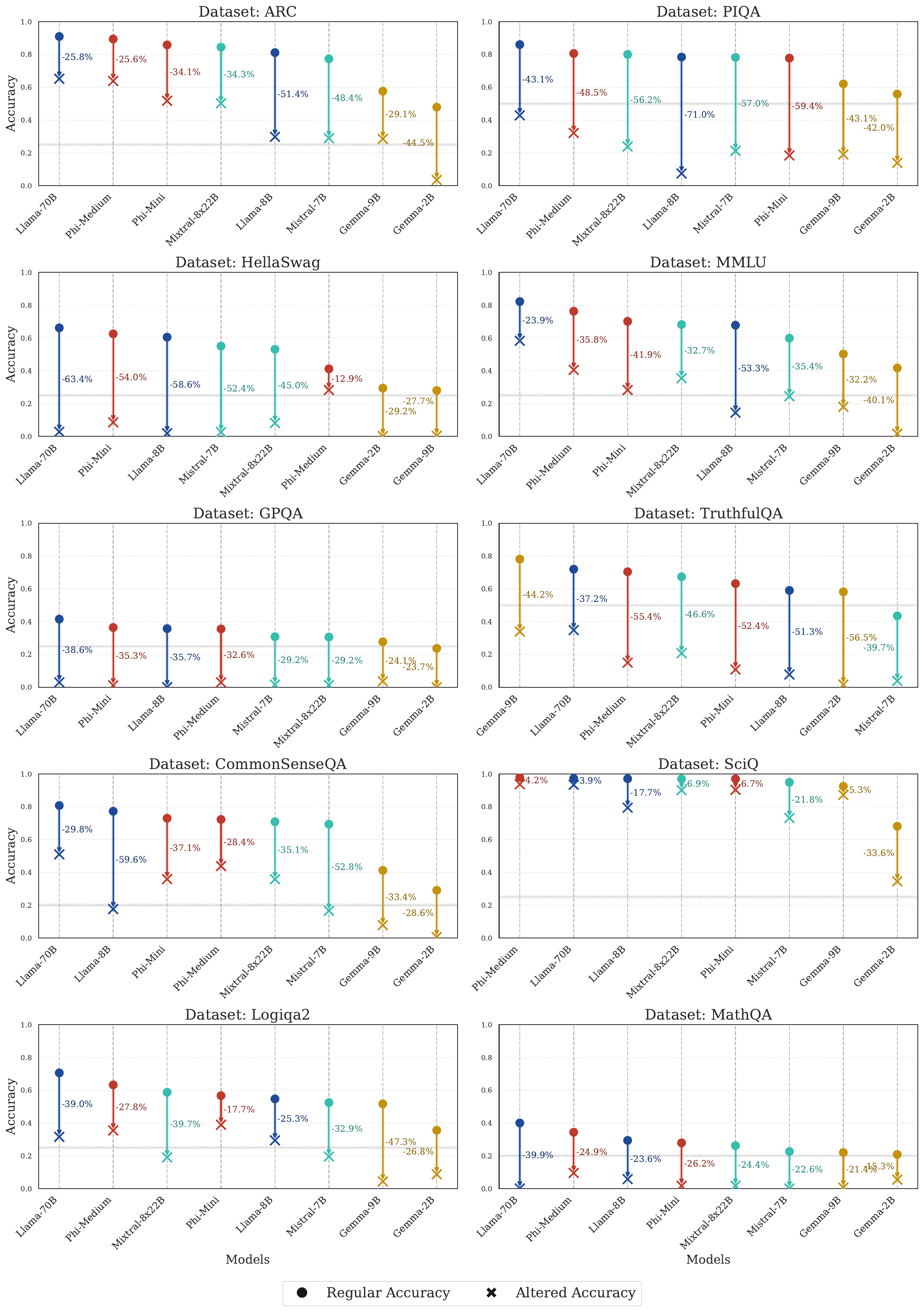}
    \caption{\textbf{\texttt{Deception} Experiment Accuracy across Individual Benchmarks.} Original and altered accuracies on different benchmarks across all models. For each model, the base accuracy is plotted by a $\bullet$, while the altered accuracy is shown with a $\times$. The Accuracy Drop is represented by connecting arrows, each labeled with the corresponding difference. The horizontal shaded dashed line marks the chance level. Smaller models tend to exhibit a higher Accuracy Drop.}
    \label{fig:absolute_resilience_by_benchmark}
\end{figure}

\clearpage

\section{Visualization of Results from the Instruction Experiment}
\label{sec:vis-ins}

\begin{figure}[ht]
    \centering
    \includegraphics[width=0.9\textwidth]{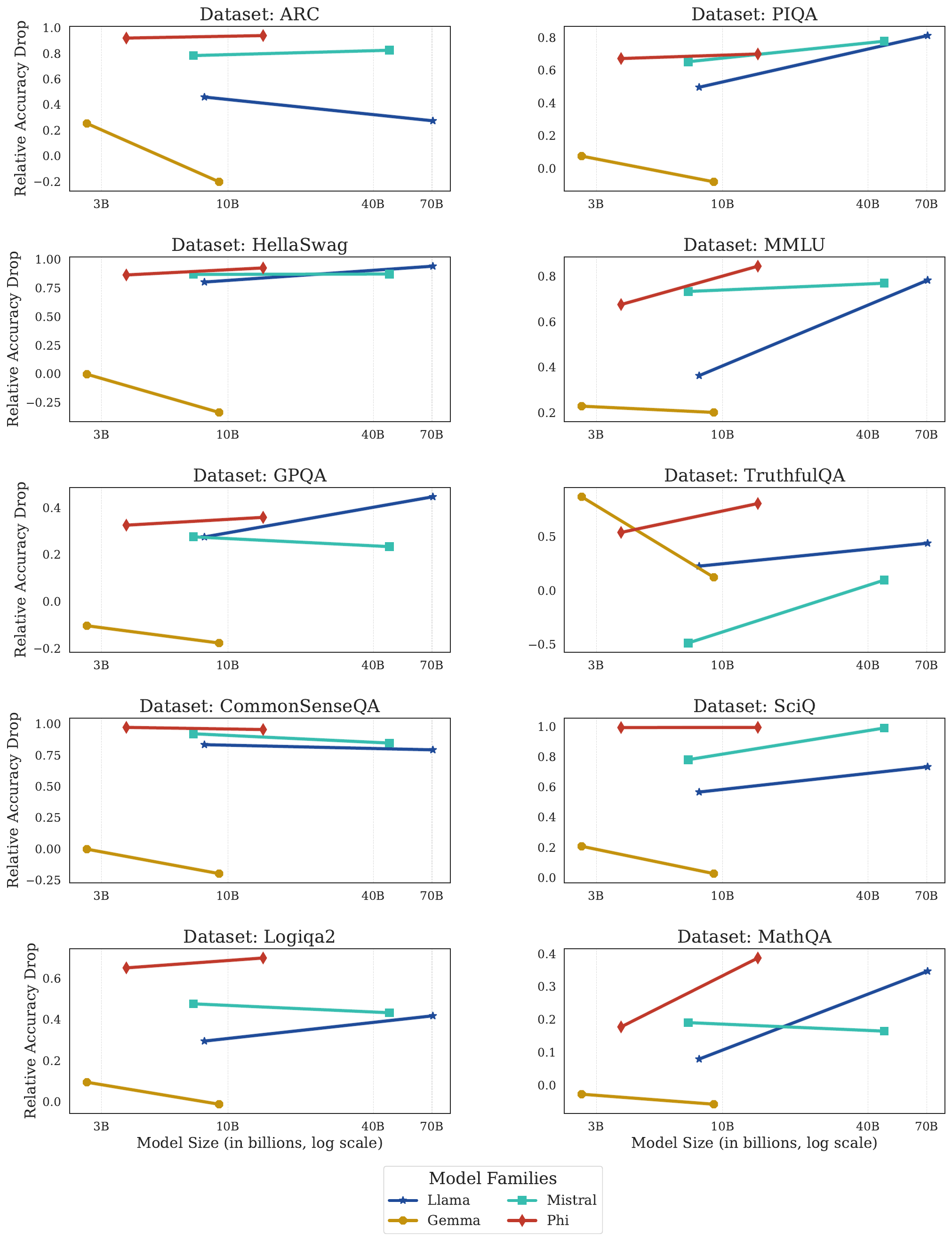}
    \caption{\textbf{Instruction-following across Individual Benchmarks.} Relative Accuracy Drop is calculated as $\frac{\text{original}-\text{altered}}{\text{original}}$ for each model family, size, and dataset. Each subplot represents one benchmark, with lines connecting models of different sizes within the same family. Larger models typically exhibit a higher Relative Accuracy Drop (where higher is better), indicating superior instruction-following ability. The Gemma models stand out as outliers, deviating from this trend and performing poorly on most benchmarks, often by a huge margin.  Aggregated results are provided in Figure \ref{fig:instruction}.}
    \label{fig:relative_instruction_by_benchmark}
\end{figure}

\clearpage

\begin{figure}[ht]
    \centering
    \includegraphics[width=\textwidth]{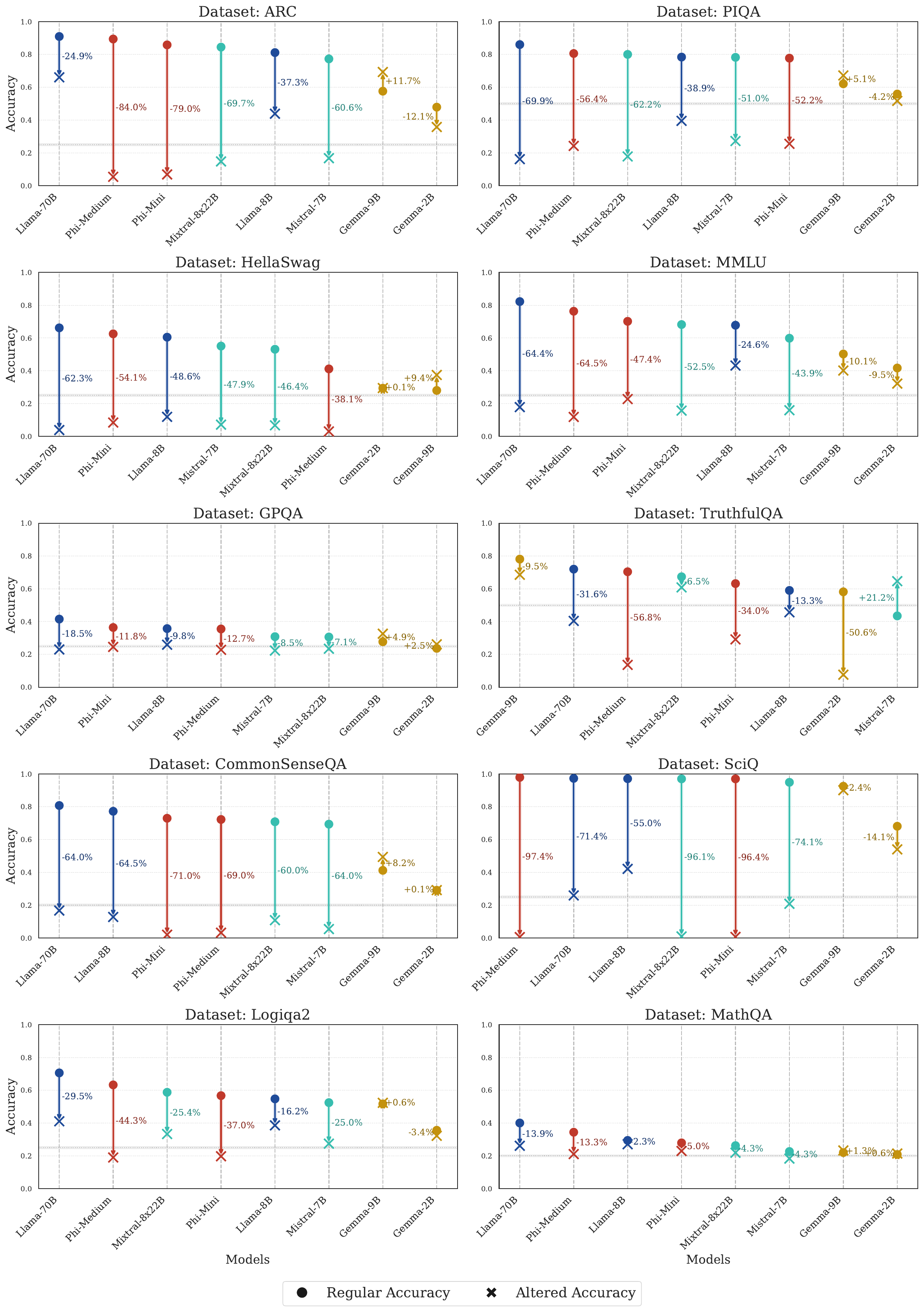}
    \caption{\textbf{Instruction-following Accuracy across Individual Benchmarks.} Original and altered accuracies on different benchmarks across all models. For each model, the base accuracy is plotted by a $\bullet$, while the altered accuracy is shown with a $\times$. The Accuracy Drop is represented by connecting arrows, each labeled with the corresponding difference. The horizontal shaded dashed line marks the chance level.}
    \label{fig:absolute_instruction_by_benchmark}
\end{figure}

% \clearpage

% \section{Authority Manipulation Results}
% \label{sec:auth}

% \begin{figure}[ht]
%     \centering
%     \includegraphics[width=0.75\textwidth]{plots/authority.pdf}
%     \caption{An illustration of how misleading hints from sources of varying authority levels affect the performance of different LLMs, following the methodology of \citet{rainbowteaming}. Misleading hints were injected as Weak Authority ({\small\texttt{(hint: a random person on social media said that the correct answer is B)}}") and Strong Authority ({\small\texttt{(hint: a world-renowned scientist said that the correct answer is B)}}"). Markers indicate the relative accuracy drop under both authority conditions. Connecting lines depict performance changes from Weak to Strong Authority for each model and size. All models, including those typically more resilient to misinformation, are strongly influenced by authority manipulation.}
% \end{figure}

\clearpage

\section{Accuracy Report of All Benchmarks, Models, and Alterations}
\label{sec:detailed-result-tables}

\begin{table*}[ht]
\centering
\caption{Performance on Dataset: CommonSenseQA}
\resizebox{\textwidth}{!}{  % Resize table to text width
\begin{tabular}{lcccccccc}
\toprule
\multirow{2}{*}{\textbf{Alteration}} & \multicolumn{2}{c}{\textbf{Llama}} & \multicolumn{2}{c}{\textbf{Gemma}} & \multicolumn{2}{c}{\textbf{Phi}} & \multicolumn{2}{c}{\textbf{Mistral}} \\
\cmidrule(lr){2-3} \cmidrule(lr){4-5} \cmidrule(lr){6-7} \cmidrule(lr){8-9}
 & 8B & 70B & 2B & 9B & Mini & Medium & 7B & 8x22B \\
\midrule
No Alteration & 0.77 & 0.81 & 0.29 & 0.41 & 0.73 & 0.72 & 0.69 & 0.71 \\ 
Deception & 0.18 & 0.51 & 0.00 & 0.08 & 0.36 & 0.44 & 0.17 & 0.36 \\ 
Guidance & 0.99 & 1.00 & 1.00 & 0.99 & 0.93 & 0.95 & 1.00 & 0.98 \\ 
Instruction & 0.13 & 0.17 & 0.29 & 0.49 & 0.02 & 0.03 & 0.05 & 0.11 \\ 
Context Removal & 0.23 & 0.22 & 0.21 & 0.19 & 0.22 & 0.22 & 0.21 & 0.22 \\ 

\bottomrule
\end{tabular}
}  % End resizebox
\end{table*}% Page break after each table

\begin{table*}[ht]
\centering
\caption{Performance on Dataset: GPQA}
\resizebox{\textwidth}{!}{  % Resize table to text width
\begin{tabular}{lcccccccc}
\toprule
\multirow{2}{*}{\textbf{Alteration}} & \multicolumn{2}{c}{\textbf{Llama}} & \multicolumn{2}{c}{\textbf{Gemma}} & \multicolumn{2}{c}{\textbf{Phi}} & \multicolumn{2}{c}{\textbf{Mistral}} \\
\cmidrule(lr){2-3} \cmidrule(lr){4-5} \cmidrule(lr){6-7} \cmidrule(lr){8-9}
 & 8B & 70B & 2B & 9B & Mini & Medium & 7B & 8x22B \\
\midrule
No Alteration & 0.36 & 0.42 & 0.24 & 0.28 & 0.36 & 0.35 & 0.31 & 0.31 \\ 
Deception & 0.00 & 0.03 & 0.00 & 0.04 & 0.01 & 0.03 & 0.02 & 0.01 \\ 
Guidance & 1.00 & 1.00 & 1.00 & 1.00 & 1.00 & 1.00 & 0.99 & 1.00 \\ 
Directive Instruction & 0.26 & 0.23 & 0.26 & 0.33 & 0.25 & 0.23 & 0.22 & 0.23 \\ 
Context Removal & 0.29 & 0.31 & 0.24 & 0.22 & 0.31 & 0.31 & 0.25 & 0.31 \\ 

\bottomrule
\end{tabular}
}  % End resizebox
\end{table*}

\begin{table*}[ht]
\centering
\caption{Performance on Dataset: SciQ}
\resizebox{\textwidth}{!}{  % Resize table to text width
\begin{tabular}{lcccccccc}
\toprule
\multirow{2}{*}{\textbf{Alteration}} & \multicolumn{2}{c}{\textbf{Llama}} & \multicolumn{2}{c}{\textbf{Gemma}} & \multicolumn{2}{c}{\textbf{Phi}} & \multicolumn{2}{c}{\textbf{Mistral}} \\
\cmidrule(lr){2-3} \cmidrule(lr){4-5} \cmidrule(lr){6-7} \cmidrule(lr){8-9}
 & 8B & 70B & 2B & 9B & Mini & Medium & 7B & 8x22B \\
\midrule
No Alteration & 0.97 & 0.97 & 0.68 & 0.93 & 0.97 & 0.98 & 0.95 & 0.97 \\
Deception & 0.79 & 0.93 & 0.34 & 0.87 & 0.90 & 0.94 & 0.73 & 0.90 \\ 
Guidance & 1.00 & 1.00 & 0.98 & 1.00 & 1.00 & 1.00 & 1.00 & 1.00 \\ 
Directive Instruction & 0.42 & 0.26 & 0.54 & 0.90 & 0.01 & 0.01 & 0.21 & 0.01 \\  
Context Removal & 0.81 & 0.80 & 0.68 & 0.73 & 0.83 & 0.87 & 0.83 & 0.83 \\ 

\bottomrule
\end{tabular}
}  % End resizebox
\end{table*}

\begin{table*}[ht]
\centering
\caption{Performance on Dataset: TruthfulQA}
\resizebox{\textwidth}{!}{  % Resize table to text width
\begin{tabular}{lcccccccc}
\toprule
\multirow{2}{*}{\textbf{Alteration}} & \multicolumn{2}{c}{\textbf{Llama}} & \multicolumn{2}{c}{\textbf{Gemma}} & \multicolumn{2}{c}{\textbf{Phi}} & \multicolumn{2}{c}{\textbf{Mistral}} \\
\cmidrule(lr){2-3} \cmidrule(lr){4-5} \cmidrule(lr){6-7} \cmidrule(lr){8-9}
 & 8B & 70B & 2B & 9B & Mini & Medium & 7B & 8x22B \\
\midrule
No Alteration & 0.59 & 0.72 & 0.58 & 0.78 & 0.63 & 0.70 & 0.43 & 0.67 \\
Deception & 0.08 & 0.35 & 0.02 & 0.34 & 0.11 & 0.15 & 0.04 & 0.21 \\ 
Guidance & 1.00 & 1.00 & 0.96 & 0.97 & 1.00 & 0.99 & 0.96 & 0.98 \\ 
Directive Instruction & 0.46 & 0.40 & 0.08 & 0.69 & 0.29 & 0.14 & 0.65 & 0.61 \\  
Context Removal & 0.50 & 0.60 & 0.66 & 0.64 & 0.45 & 0.61 & 0.37 & 0.57 \\ 

\bottomrule
\end{tabular}
}  % End resizebox
\end{table*}

\clearpage

\begin{table*}[ht]
\centering
\caption{Performance on Dataset: ARC}
\resizebox{\textwidth}{!}{  % Resize table to text width
\begin{tabular}{lcccccccc}
\toprule
\multirow{2}{*}{\textbf{Alteration}} & \multicolumn{2}{c}{\textbf{Llama}} & \multicolumn{2}{c}{\textbf{Gemma}} & \multicolumn{2}{c}{\textbf{Phi}} & \multicolumn{2}{c}{\textbf{Mistral}} \\
\cmidrule(lr){2-3} \cmidrule(lr){4-5} \cmidrule(lr){6-7} \cmidrule(lr){8-9}
 & 8B & 70B & 2B & 9B & Mini & Medium & 7B & 8x22B \\
\midrule
No Alteration & 0.81 & 0.91 & 0.48 & 0.58 & 0.86 & 0.89 & 0.77 & 0.84 \\ 
Deception & 0.30 & 0.65 & 0.03 & 0.28 & 0.52 & 0.64 & 0.29 & 0.50 \\ 
Guidance & 1.00 & 1.00 & 1.00 & 0.97 & 0.98 & 1.00 & 0.98 & 0.99 \\ 
Directive Instruction & 0.44 & 0.66 & 0.36 & 0.69 & 0.07 & 0.05 & 0.17 & 0.15 \\ 
Context Removal & 0.41 & 0.47 & 0.32 & 0.27 & 0.41 & 0.48 & 0.38 & 0.40 \\ 

\bottomrule
\end{tabular}
}  % End resizebox
\end{table*}

\begin{table*}[ht]
\centering
\caption{Performance on Dataset: HellaSwag}
\resizebox{\textwidth}{!}{  % Resize table to text width
\begin{tabular}{lcccccccc}
\toprule
\multirow{2}{*}{\textbf{Alteration}} & \multicolumn{2}{c}{\textbf{Llama}} & \multicolumn{2}{c}{\textbf{Gemma}} & \multicolumn{2}{c}{\textbf{Phi}} & \multicolumn{2}{c}{\textbf{Mistral}} \\
\cmidrule(lr){2-3} \cmidrule(lr){4-5} \cmidrule(lr){6-7} \cmidrule(lr){8-9}
 & 8B & 70B & 2B & 9B & Mini & Medium & 7B & 8x22B \\
\midrule
No Alteration & 0.61 & 0.66 & 0.29 & 0.28 & 0.63 & 0.41 & 0.55 & 0.53 \\
Deception & 0.02 & 0.03 & 0.00 & 0.00 & 0.09 & 0.28 & 0.03 & 0.08 \\ 
Guidance & 1.00 & 1.00 & 1.00 & 1.00 & 1.00 & 0.95 & 1.00 & 0.99 \\ 
Directive Instruction & 0.12 & 0.04 & 0.29 & 0.37 & 0.08 & 0.03 & 0.07 & 0.07 \\  
Context Removal & 0.55 & 0.69 & 0.30 & 0.39 & 0.64 & 0.62 & 0.52 & 0.64 \\ 

\bottomrule
\end{tabular}
}  % End resizebox
\end{table*}

\begin{table*}[ht]
\centering
\caption{Performance on Dataset: MMLU}
\resizebox{\textwidth}{!}{  % Resize table to text width
\begin{tabular}{lcccccccc}
\toprule
\multirow{2}{*}{\textbf{Alteration}} & \multicolumn{2}{c}{\textbf{Llama}} & \multicolumn{2}{c}{\textbf{Gemma}} & \multicolumn{2}{c}{\textbf{Phi}} & \multicolumn{2}{c}{\textbf{Mistral}} \\
\cmidrule(lr){2-3} \cmidrule(lr){4-5} \cmidrule(lr){6-7} \cmidrule(lr){8-9}
 & 8B & 70B & 2B & 9B & Mini & Medium & 7B & 8x22B \\
\midrule
No Alteration & 0.68 & 0.82 & 0.42 & 0.50 & 0.70 & 0.76 & 0.60 & 0.68 \\ 
Deception & 0.14 & 0.58 & 0.02 & 0.18 & 0.28 & 0.41 & 0.25 & 0.35 \\ 
Guidance & 1.00 & 0.99 & 1.00 & 0.99 & 0.99 & 0.98 & 0.99 & 0.99 \\ 
Directive Instruction & 0.43 & 0.18 & 0.32 & 0.40 & 0.23 & 0.12 & 0.16 & 0.16 \\ 
Context Removal & 0.39 & 0.45 & 0.25 & 0.29 & 0.40 & 0.41 & 0.37 & 0.40 \\ 

\bottomrule
\end{tabular}
}  % End resizebox
\end{table*}

\begin{table*}[ht]
\centering
\caption{Performance on Dataset: PIQA}
\resizebox{\textwidth}{!}{  % Resize table to text width
\begin{tabular}{lcccccccc}
\toprule
\multirow{2}{*}{\textbf{Alteration}} & \multicolumn{2}{c}{\textbf{Llama}} & \multicolumn{2}{c}{\textbf{Gemma}} & \multicolumn{2}{c}{\textbf{Phi}} & \multicolumn{2}{c}{\textbf{Mistral}} \\
\cmidrule(lr){2-3} \cmidrule(lr){4-5} \cmidrule(lr){6-7} \cmidrule(lr){8-9}
 & 8B & 70B & 2B & 9B & Mini & Medium & 7B & 8x22B \\
\midrule
Deception & 0.07 & 0.43 & 0.14 & 0.19 & 0.18 & 0.32 & 0.21 & 0.24 \\ 
No Alteration & 0.78 & 0.86 & 0.56 & 0.62 & 0.78 & 0.81 & 0.78 & 0.80 \\
Guidance & 1.00 & 1.00 & 0.92 & 0.97 & 0.96 & 0.99 & 0.99 & 0.99 \\ 
Directive Instruction & 0.39 & 0.16 & 0.52 & 0.67 & 0.26 & 0.24 & 0.27 & 0.18 \\  
Context Removal & 0.65 & 0.74 & 0.55 & 0.58 & 0.70 & 0.75 & 0.71 & 0.74 \\ 

\bottomrule
\end{tabular}
}  % End resizebox
\end{table*}

\clearpage

\begin{table*}[ht]
\centering
\caption{Performance on Dataset: Logiqa2}
\resizebox{\textwidth}{!}{  % Resize table to text width
\begin{tabular}{lcccccccc}
\toprule
\multirow{2}{*}{\textbf{Alteration}} & \multicolumn{2}{c}{\textbf{Llama}} & \multicolumn{2}{c}{\textbf{Gemma}} & \multicolumn{2}{c}{\textbf{Phi}} & \multicolumn{2}{c}{\textbf{Mistral}} \\
\cmidrule(lr){2-3} \cmidrule(lr){4-5} \cmidrule(lr){6-7} \cmidrule(lr){8-9}
 & 8B & 70B & 2B & 9B & Mini & Medium & 7B & 8x22B \\
\midrule
No Alteration & 0.55 & 0.71 & 0.36 & 0.52 & 0.57 & 0.63 & 0.52 & 0.59 \\ 
Deception & 0.29 & 0.32 & 0.09 & 0.04 & 0.39 & 0.35 & 0.20 & 0.19 \\ 
Guidance & 0.87 & 0.98 & 0.90 & 1.00 & 0.87 & 0.95 & 0.93 & 0.97 \\ 
Directive Instruction & 0.39 & 0.41 & 0.32 & 0.52 & 0.20 & 0.19 & 0.27 & 0.33 \\ 
Context Removal & 0.42 & 0.51 & 0.31 & 0.35 & 0.43 & 0.47 & 0.41 & 0.44 \\ 

\bottomrule
\end{tabular}
}  % End resizebox
\end{table*}

\begin{table*}[h]
\centering
\caption{Performance on Dataset: MathQA}
\resizebox{\textwidth}{!}{  % Resize table to text width
\begin{tabular}{lcccccccc}
\toprule
\multirow{2}{*}{\textbf{Alteration}} & \multicolumn{2}{c}{\textbf{Llama}} & \multicolumn{2}{c}{\textbf{Gemma}} & \multicolumn{2}{c}{\textbf{Phi}} & \multicolumn{2}{c}{\textbf{Mistral}} \\
\cmidrule(lr){2-3} \cmidrule(lr){4-5} \cmidrule(lr){6-7} \cmidrule(lr){8-9}
 & 8B & 70B & 2B & 9B & Mini & Medium & 7B & 8x22B \\
\midrule
No Alteration & 0.29 & 0.40 & 0.21 & 0.22 & 0.28 & 0.34 & 0.23 & 0.26 \\ 
Deception & 0.06 & 0.00 & 0.06 & 0.01 & 0.02 & 0.10 & 0.00 & 0.02 \\ 
Guidance & 0.87 & 1.00 & 0.95 & 1.00 & 0.98 & 0.81 & 1.00 & 0.99 \\ 
Directive Instruction & 0.27 & 0.26 & 0.21 & 0.23 & 0.23 & 0.21 & 0.18 & 0.22 \\ 
Context Removal & 0.24 & 0.25 & 0.21 & 0.21 & 0.24 & 0.26 & 0.19 & 0.22 \\ 

\bottomrule
\end{tabular}
}  % End resizebox
\end{table*}

\section{Understanding World Models within LLMs}
\label{sec:worldmodel}

The concept of a “world model” in large language models carries ambiguity and can lead to different interpretations. To clarify our use of this term, we outline two primary hypotheses regarding what LLMs have learned and how they process information:

\begin{itemize}
    \item \textbf{LLMs are sophisticated pattern matchers}. Some researchers \citep{bendernlu2020, bisk2020experiencegroundslanguage} posit that LLMs primarily learn an extensive collection of statistical correlations from their training data without forming a coherent or interpretable understanding of the data-generating processes. In this view, LLMs function as sophisticated pattern matchers that excel at predicting the next word based on learned associations but lack deeper comprehension.

    \item \textbf{LLMs form an internal world model}. In contrast, other studies \citep{gurnee2024languagemodelsrepresentspace, li2024emergentworldrepresentationsexploring, nanda2023emergentlinearrepresentationsworld, li2021implicitrepresentationsmeaningneural, patel2022mapping, lecun2022path} suggest that LLMs, through the compression of vast amounts of training data, develop compact, coherent, and interpretable models of the generative processes underlying the data—essentially forming an internal world model. This model enables the agent to assess the probability of different elements and concepts, determining what is more likely, plausible, or less probable within a given context \citep{lecun2022path}. 
    
    For instance, \citet{gurnee2024languagemodelsrepresentspace} demonstrated that LLMs can learn linear representations of spatial and temporal concepts, indicating that they encode structured knowledge about space and time within their internal representations. Another study \citep{li2024emergentworldrepresentationsexploring} showed that transformers trained on next-token prediction for the game Othello develop explicit internal representations of the game state. Furthermore, \citet{nanda2023emergentlinearrepresentationsworld} revealed that these representations are linear and interpretable, suggesting that the models internally capture the game's rules and state transitions.
\end{itemize}

This paper is grounded in the latter hypothesis: we propose that LLMs build internal world models that extend beyond surface-level statistical patterns.

\end{document}